\documentclass[journal]{IEEEtran}

\usepackage[mathscr]{eucal}
\usepackage[cmex10]{amsmath}
\usepackage{epsfig,epsf,psfrag}
\usepackage{amssymb,amsmath,amsthm,amsfonts,latexsym}
\usepackage{amsmath,graphicx,bm,xcolor,url}
\usepackage[caption=false]{subfig}
\usepackage{fixltx2e}
\usepackage{array}
\usepackage{verbatim}
\usepackage{bm}
\usepackage{algorithmic, cite}
\usepackage{algorithm}
\usepackage{verbatim}
\usepackage{textcomp}
\usepackage{mathrsfs}
\usepackage{enumerate}
\usepackage{epstopdf}

\catcode`~=11 \def\UrlSpecials{\do\~{\kern -.15em\lower .7ex\hbox{~}\kern .04em}} \catcode`~=13

\allowdisplaybreaks[3]

\newcommand{\calC}{\mathcal{C}}

\newcommand{\calI}{\mathcal{I}}

\newcommand{\calL}{\mathcal{L}}
\newcommand{\calM}{\mathcal{M}}
\newcommand{\calN}{\mathcal{N}}
\newcommand{\calO}{\mathcal{O}}

\newcommand{\calR}{\mathcal{R}}
\newcommand{\calS}{\mathcal{S}}
\newcommand{\calT}{\mathcal{T}}

\newcommand{\ba}{\mathbf{a}}
\newcommand{\bA}{\mathbf{A}}

\newcommand{\bc}{\mathbf{c}}
\newcommand{\bC}{\mathbf{C}}
\newcommand{\bd}{\mathbf{d}}
\newcommand{\bD}{\mathbf{D}}

\newcommand{\bE}{\mathbf{E}}

\newcommand{\bg}{\mathbf{g}}
\newcommand{\bG}{\mathbf{G}}
\newcommand{\bh}{\mathbf{h}}
\newcommand{\bH}{\mathbf{H}}

\newcommand{\bI}{\mathbf{I}}

\newcommand{\bM}{\mathbf{M}}

\newcommand{\bq}{\mathbf{q}}
\newcommand{\bQ}{\mathbf{Q}}

\newcommand{\bT}{\mathbf{T}}

\newcommand{\bU}{\mathbf{U}}

\newcommand{\bV}{\mathbf{V}}
\newcommand{\bw}{\mathbf{w}}
\newcommand{\bW}{\mathbf{W}}
\newcommand{\bx}{\mathbf{x}}
\newcommand{\bX}{\mathbf{X}}
\newcommand{\by}{\mathbf{y}}
\newcommand{\bY}{\mathbf{Y}}
\newcommand{\bz}{\mathbf{z}}
\newcommand{\bZ}{\mathbf{Z}}

\DeclareMathAlphabet{\mathbsf}{OT1}{cmss}{bx}{n}
\DeclareMathAlphabet{\mathssf}{OT1}{cmss}{m}{sl}

\DeclareSymbolFont{bsfletters}{OT1}{cmss}{bx}{n}
\DeclareSymbolFont{ssfletters}{OT1}{cmss}{m}{n}
\DeclareMathSymbol{\bsfGamma}{0}{bsfletters}{'000}
\DeclareMathSymbol{\ssfGamma}{0}{ssfletters}{'000}
\DeclareMathSymbol{\bsfDelta}{0}{bsfletters}{'001}
\DeclareMathSymbol{\ssfDelta}{0}{ssfletters}{'001}
\DeclareMathSymbol{\bsfTheta}{0}{bsfletters}{'002}
\DeclareMathSymbol{\ssfTheta}{0}{ssfletters}{'002}
\DeclareMathSymbol{\bsfLambda}{0}{bsfletters}{'003}
\DeclareMathSymbol{\ssfLambda}{0}{ssfletters}{'003}
\DeclareMathSymbol{\bsfXi}{0}{bsfletters}{'004}
\DeclareMathSymbol{\ssfXi}{0}{ssfletters}{'004}
\DeclareMathSymbol{\bsfPi}{0}{bsfletters}{'005}
\DeclareMathSymbol{\ssfPi}{0}{ssfletters}{'005}
\DeclareMathSymbol{\bsfSigma}{0}{bsfletters}{'006}
\DeclareMathSymbol{\ssfSigma}{0}{ssfletters}{'006}
\DeclareMathSymbol{\bsfUpsilon}{0}{bsfletters}{'007}
\DeclareMathSymbol{\ssfUpsilon}{0}{ssfletters}{'007}
\DeclareMathSymbol{\bsfPhi}{0}{bsfletters}{'010}
\DeclareMathSymbol{\ssfPhi}{0}{ssfletters}{'010}
\DeclareMathSymbol{\bsfPsi}{0}{bsfletters}{'011}
\DeclareMathSymbol{\ssfPsi}{0}{ssfletters}{'011}
\DeclareMathSymbol{\bsfOmega}{0}{bsfletters}{'012}
\DeclareMathSymbol{\ssfOmega}{0}{ssfletters}{'012}

\DeclareMathOperator{\sgn}{sgn}

\DeclareMathOperator{\col}{col}  

\newtheorem{definition}{Definition}

\newtheorem{data model}{Data Model}

\newcommand{\qednew}{\nobreak \ifvmode \relax \else
      \ifdim\lastskip<1.5em \hskip-\lastskip
      \hskip1.5em plus0em minus0.5em \fi \nobreak
      \vrule height0.75em width0.5em depth0.25em\fi}

\DeclareMathOperator{\innov}{innov}

\begin{document}
\title{Subspace Clustering via Optimal Direction Search}

\author{Mostafa~Rahmani, \IEEEmembership{Student Member,~IEEE} and George~K.~Atia,~\IEEEmembership{Member,~IEEE} 
\thanks{This work was supported by NSF CAREER Award CCF-1552497 and NSF Grant CCF-1320547.

The authors are with the Department of Electrical and Computer Engineering, University of Central Florida, Orlando, FL 32816 USA (e-mails: mostafa@knights.ucf.edu, george.atia@ucf.edu).}
}

\markboth{}%
{Shell \MakeLowercase{\textit{et al.}}: Bare Demo of IEEEtran.cls for Journals}
\maketitle

\begin{abstract}
This paper presents a new spectral-clustering-based approach to the subspace clustering problem.
Underpinning the proposed method is a convex program for optimal direction search, which for each data point $\bd$, finds an optimal direction in the span of the data that has minimum projection on the other data points and non-vanishing projection on $\bd$.
The obtained directions are subsequently leveraged to identify a neighborhood set for each data point. An Alternating Direction Method of Multipliers (ADMM) framework is provided to efficiently solve for the optimal directions. The proposed method is shown to often outperform the existing subspace clustering methods, particularly for unwieldy scenarios involving high levels of noise and close subspaces, and yields the state-of-the-art results for the problem of face clustering using subspace segmentation.
\end{abstract}

\begin{IEEEkeywords}
 Spectral Clustering, Convex Optimization, Unsupervised
Learning, Face Clustering, Innovation pursuit
\end{IEEEkeywords}

\IEEEpeerreviewmaketitle

\section{Introduction}
In many applications of signal processing and machine learning, the data can be well-approximated with low-dimensional subspaces~\cite{bishop2006pattern}. Subspace recovery methods have been instrumental in reducing dimensionality and recognizing intrinsic patterns in data. Principal Component Analysis (PCA) is a standard tool which approximates the data with a single low-dimensional subspace that has minimum distance from the data points~\cite{rahmani2015randomized,zhang2014novel,lerman2015robust}.
However, in many applications the data admits clustering structures, wherefore a union of subspaces can better model the data~\cite{vidal2011subspace}.

In the subspace clustering problem, the data points lie in a union of an unknown number of unknown linear subspaces whose dimensions are also generally unknown.
The role of a subspace segmentation algorithm then is to learn these low-dimensional subspaces and to cluster the data points to their respective subspaces. This data model has been widely applied to many modern signal processing and machine learning applications, including computer vision  \cite{vidal2011subspace,ho2003clustering}, gene expression analysis \cite{mcwilliams2014subspace,kriegel2009clustering}, and image processing \cite{yang2008unsupervised}. 
%

Many different approaches to subspace clustering were devised in related work, including statistical-based approaches \cite{yang2006robust,stat1,stat2,rnc1}, spectral clustering \cite{elhamifar2013sparse}, the algebraic-geometric approach \cite{vidal2005generalized}, the innovation pursuit approach \cite{rahmani2015innovationn}, and iterative methods \cite{bradley2000k,zhang2009median}. We refer the reader to \cite{vidal2011subspace,elhamifar2013sparse,rahmani2015innovationn} for an overview of the topic.
Much of the recent work has focused on spectral-clustering \cite{von2007tutorial} based methods \cite{dyer2013greedy,gao2015multi,elhamifar2013sparse,heckel2013robust,liu2013robust,soltanolkotabi2012geometric,wang2013provable,chen2009spectral,park2014greedy}, which all share a common structure.
Specifically, a neighborhood set for each data point is first identified to construct a similarity matrix. Subsequently, spectral clustering \cite{von2007tutorial} is applied to the similarity matrix. Spectral-clustering-based methods differ mostly in the first step.

There exists several recent spectral-clustering-based methods with superior empirical performance. SSC is a popular spectral-clustering-based algorithm, which finds a sparse representation for each data point with respect to the rest of data to construct the similarity matrix \cite{elhamifar2013sparse}.
It was shown in \cite{soltanolkotabi2012geometric} that SSC can yield exact clustering even for subspaces with intersections under
certain conditions. A different algorithm called Low-Rank Representation (LRR) \cite{liu2013robust} uses nuclear norm minimization to build the similarity matrix. In \cite{heckel2013robust}, the inner product between the data points is used as measure of similarity to find a neighborhood set for each data point.

\subsection{Summary of contributions}
This paper presents a new spectral-clustering-based subspace segmentation method dubbed Direction search based Subspace Clustering (DSC). Underlying our approach is a direction search program that associates an optimal direction with each data point. For each data point, the algorithm finds an optimal direction in the column space of the data matrix that has minimum projection on the rest of the data and non-vanishing projection on that data point.
An optimization framework is presented to find all the directions by solving one convex program. Subsequently, the similarity matrix is formed using the obtained directions. The presented numerical experiments demonstrate that DSC often outperforms existing spectral-clustering-based methods, and remarkably improves over the state-of-the-art result for the problem of face clustering using subspace segmentation. In addition, an iterative method to efficiently solve the proposed direction search optimization is provided.


\subsection{Notation and data model}
Bold-face upper-case letters are used to denote matrices and bold-face lower-case letters are used to denote vectors. For a vector $\ba$, $\| \ba \|_p$ denotes its $\ell_p$-norm. 
Given two matrices $\bA_1$ and $\bA_2$ with an equal number of rows, the matrix
$
\bA = [\bA_1 \: \: \bA_2]
$
is the matrix formed from the concatenation of $\bA_1$ and $\bA_2$. Given a vector $\ba$, $\big | \ba | $ is the vector of absolute values of the elements of $\ba$. Given a matrix $\bA$, $\ba_i$ denotes its $i^{\text{th}}$ column,
 $\| \bA \|_{1,1} = \sum_{i} \| \ba_i \|_1$, and $\| \bA \|_{1,2} = \sum_{i} \| \ba_i \|_2$,  $\col(\bA)$  its column space, and $\text{tr} (\bA)$ its trace. In addition, $\text{diag} (\bA)$ returns a vector of the diagonal elements of $\bA$.
The symbol $\oplus$ denotes the direct sum operator.

In this paper, the data is assumed to follow the subspace clustering structure expressed in the following data model.
\\
\\
\textbf{Data Model 1.} 
\textit{
The data matrix $\bD \in \mathbb{R}^{M_1 \times M_2}$ can be represented as
$\bD = [\bD_1 \: ... \: \bD_N] \bT$, where $\bT$ is an arbitrary permutation matrix. The columns of $\bD_i \in \mathbb{R}^{M_1 \times n_i}$ lie in $\mathcal{S}_i$, where $\mathcal{S}_i$ is an $r_i$-dimensional linear subspace, for $1 \leq i \leq N$, and, $\sum_{i = 1}^N n_i = M_2$. 
}

\smallbreak
\noindent
We define $\bQ \in \mathbb{R}^{M_1 \times r}$ as an orthonormal basis for $\col(\bD)$, where $r$ is the rank of $\bD$. If the data is noisy, the matrix $\bQ$ is formed using the dominant left singular vectors of $\bD$. In addition, $\bX \in \mathbb{R}^{r \times M_2}$ is defined as $\bX = \bQ^T \bD$.


\section{Direction search clustering}
The proposed approach consists of $M_2$ identical optimization problems, one per data point. The optimization problem 
\begin{eqnarray}
 \underset{\ba }{\min}
 \| \ba^T \bX \|_p \quad \text{s.t.}  \quad   \ba^T  \bx_i = 1 \:
 \label{eq:robust21}
\end{eqnarray}
corresponding to $\bd_i$, searches for a direction in the column space of the projected data $\bX$ with non-zero projection on $\bx_i$ and minimum projection on the rest of the data. 
In this paper, we use $p=1$ or $p=2$ for the $\ell_p$-norm. The linear constraint enforces the optimal point of (\ref{eq:robust21}) to have strong coherence with $\bx_i$. In practice, the data points within a subspace are mutually coherent, wherefore the optimal point of (\ref{eq:robust21}) will have large projection on other data points in the subspace containing ${\bx}_i$. Accordingly, if we sample few of the columns of $\bX$ corresponding to the elements of $| \ba^T \bX |$ with largest values, they will all lie in the subspace containing $\bx_i$. Thus, we exploit the obtained directions to construct a neighborhood set for each data point in order to construct a similarity matrix, hence the name Direction search Subspace Clustering (DSC). Algorithm 1 describes the proposed DSC method. The first step finds all the directions in one shot via solving a $r \times M_2$ convex optimization problem. The similarity matrix is formed in the second step, then in the final step the spectral clustering algorithm is applied to the similarity matrix. For more information about Steps 2.2 and 3, the reader is referred to \cite{von2007tutorial,vidal2011subspace}.
\begin{figure}[t!]
	\centering
    \includegraphics[width=0.5\textwidth]{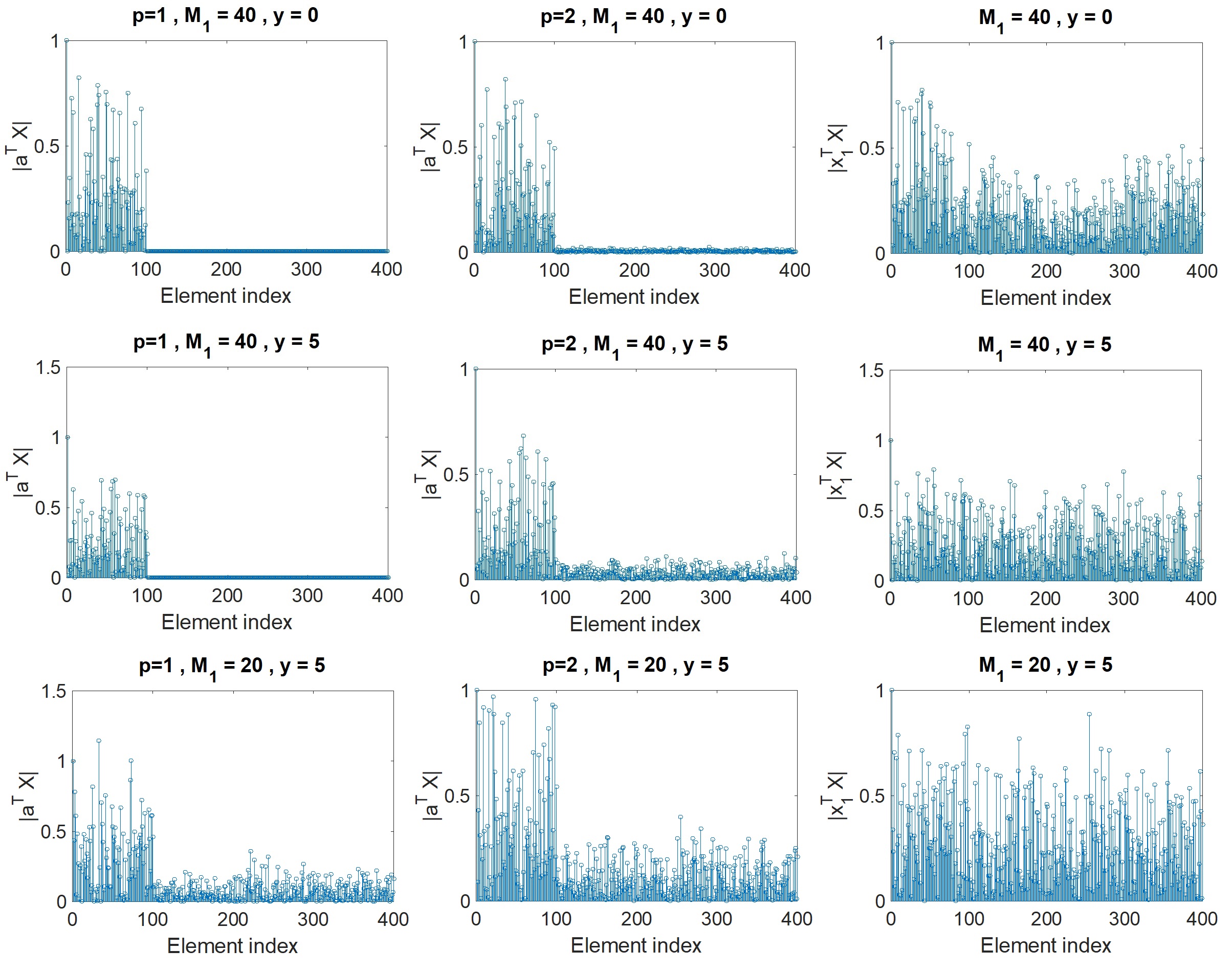}
    \vspace{-.3in}
    \caption{Measures of similarity $|\ba^T \bX|$ and $|\bx_1^T\bX|$ adopted by DSC and TSC to identify a neighborhood set for the first data point. First row $M_1 = 40$ and $y=0$, second row $M_1 = 40$ and $y=5$, third row $M_1 = 20$ and $y=5$.}
    \label{fig:example}
\end{figure}
\begin{algorithm}
\caption{Direction search based Subspace Clustering (DSC)}
{\footnotesize
\textbf{Initialization:} Set $g$ equal to the cardinality of a neighborhood set. Set $\bW \in \mathbb{R}^{M_2 \times M_2}$ equal to a zero matrix and set $p$ equal to 1 or 2.

Normalize the $\ell_2$-norm of the columns of $\bD$ (i.e., set $\bd_i$ equal to $\bd_i/\| \bd_i \|_2$). Form matrix $\bX = \bQ^T \bD$.

\smallbreak
\textbf{1. Define $\bA^{*}$} as the optimal points of
\begin{eqnarray}
\begin{aligned}
& \underset{\bA}{\min}\:   \| \bX^T \bA \|_{1,p}  \quad \text{subject to} \quad  \text{diag}( \bA^T \bX) = \textbf{1} \:,
\end{aligned}
\label{eq:main_op_p}
\end{eqnarray}
where $\textbf{1} \in \mathbb{R}^{M_2 \times 1}$ is the vector of all ones.

\smallbreak
\textbf{2. For} $i=1$ to $M_2$\\
\textbf{2.1} Set $\calI$ equal to the index set of the largest $g$ elements of $\left| \ba_i^{*} \bX  \right|$. 
\\
\textbf{2.2} $\bw_{\calI}^i = \text{exp}(-2*\text{acos}({\bx}_i^T {\bX}_{\calI}))$, where $\text{acos}$ and $\text{exp}$ are the element-wise inverse cosine and exponential functions, respectively, $\bX_{\calI}$ the columns of $\bX$ indexed by $\calI$, $\bw^i$ the $i^{\text{th}}$ row of $\bW$, and $\bw_{\calI}^i$ the elements of $\bw^i$ indexed by $\calI$. \\
\textbf{2. End For}

\smallbreak

\textbf{3.} Apply spectral clustering to the matrix $\bW+\bW^T$.
}
\end{algorithm}

\noindent\textit{Sparse regularization:}
If the data matrix is low rank, each vector in $\col(\bQ)$  can
be represented as a sparse combination of the columns of $\bD$. For such setting, we can rewrite (\ref{eq:main_op_p}) as
\begin{eqnarray}
\begin{aligned}
& \underset{\bA , \bZ}{\min}
& & \|  \bA^T \bX \|_{1,p} + \gamma \|\bZ\|_1  \\
& \text{subject to}
& & \bA = \bX \: \bZ \\
& & & \text{diag} \left( \bA^T \bX\right)  = \textbf{1} \:,
\end{aligned}
\label{eq: sparse rep}
\end{eqnarray}
where  $\bZ \in \mathbb{R}^{M_2 \times M_2}$. The sparse representation can further enhance the robustness of the proposed approach to noise. The singular vectors corresponding to the noise component do not admit sparse representations in the data -- that is, are normally obtained through linear combinations of a large number of data points. Thus, enforcing a sparse representation for the optimal direction averts a solution for
\begin{eqnarray}
\begin{aligned}
& \underset{\ba , \bz}{\min}
& & \|  \ba^T \bX \|_{p} + \gamma \|\bz\|_1  \\
& \text{subject to}
& & \ba = \bX  \bz \:\:,\: \: \ba^T \bx_i = 1\:,
\end{aligned}
\end{eqnarray}
that
lies in close proximity with the noise singular vectors.
\subsection{Connection and contrast to TSC and iPursuit}
We point out some similarities and fundamental differences between DSC and some of the more related approaches.
DSC and TSC bear some resemblance from a structural standpoint, yet are conceptually very different concerning how data similarity is viewed and measured, and thus how neighborhoods are constructed. Specifically, underlying DSC is the convex program (\ref{eq:main_op_p}) whereby optimal directions are obtained in Step 1 to be used in Step 2.1 of Algorithm 1 to construct the similarity matrix.
This is fundamentally different from the thresholding-based subspace clustering (TSC) algorithm \cite{heckel2013robust}, which uses the data points themselves as directions. Thus, in TSC the equivalent of set $\calI$ is formed from the indices of the largest elements of $|\bd_i^T \bD|$. Hence, the performance of TSC greatly declines when the subspaces are in close proximity.

As an example, suppose the columns of $\bD = [\bD_1 \: \: \bD_2 \: \: \bD_3 \:\:\bD_4]$ lie in the union of 4 10-dimensional subspaces $\{ \calS_i \}_{i=1}^4$, each with a 100 data points, where $\calS_i = \calM \oplus \calR_i$. $\calM$ is a random $y$-dimensional subspace, and $\{ \calR_i \}_{i=1}^4$ are random $(10-y)$-dimensional subspaces. Thus, the dimension of the intersections between the subspaces is equal to $y$ with high probability. We solve (\ref{eq:robust21}) with $i=1$.
The first two columns of Fig. \ref{fig:example} illustrate the values of $|\ba^T \bX|$ for $p=1$ and $p=2$ adopted by DSC as a measure of similarity to build the neighborhood set of the first data point, and the last column displays the values of $|\bx_1^T \bX |$ adopted by TSC. In the first row $M_1 = 40$ and $y=0$, corresponding to independent subspaces that do not intersect. For, the second row $y=5$ (i.e., closer subspaces) and in the last row $M_1 = 20$ and $y=5$.
As desired, the largest values of $|\ba^T \bX|$ used to form the set $\cal I$ in Step 2.1 consistently correspond to the first subspace. 
When $y=0$, the subspaces are not very close to each other and TSC can build a correct neighborhood for $\bd_1$ since the data columns corresponding to the largest elements of $|\bx_1^T \bX |$ all lie in the same subspace $\calS_1$.
However, in the second and third row where $y=5$, TSC cannot form a proper neighborhood as the data points corresponding to the largest elements of $|\bx_1^T \bX|$ do not lie in the same cluster.
 Despite the close proximity of the subspaces, (\ref{eq:robust21}) finds a direction in the data span that is strongly coherent with the first subspace and has small projection on the other subspaces. 
This feature notably empowers DSC to distinguish the data clusters.

In \cite{rahmani2015innovationn}, we developed an iterative subspace clustering approach termed iPursuit (short for innovation pursuit). Akin to DSC, iPursuit leverages some direction search module for subspace identification, albeit the approach is very different. To describe the connection to, and difference from, DSC we need the following definition.
%
\begin{definition}(Innovation subspace)
Suppose $\bD$ follows Data Model 1, $\bV_i$ is an orthonormal basis for $\calS_i$,
$\bC_i$ is an orthonormal basis for ${\overset{N}{\underset{k = 1 \atop k \neq i}{\oplus}}} \calS_k$, and
 $\calS_i \nsubseteq {\overset{N}{\underset{k = 1 \atop k \neq i}{\oplus}}} \calS_k$ (i.e., $\calS_i$ does not lie completely in the direct sum of the other subspaces). Then, the innovation subspace corresponding to $\calS_i$, denoted $\innov(\calS_i)$, is defined as the linear subspace in $\col(\bD)$ spanned by the columns of $ (\bI - \bC_i \bC_i^T) \bV_i$.
 \label{def:innv_subs}
\end{definition}

In \cite{rahmani2015innovationn}, it was shown that if $\calS_1\nsubseteq\oplus_{i=2}^N \calS_i$, and  $\bq \in \mathbb{R}^{M_1}$ is sufficiently close to $\innov(\calS_1)$, then the optimal point of
\begin{eqnarray}
 \underset{\ba }{\min}
 \| \ba^T \bX \|_1 \quad \text{s.t.}  \quad   \ba^T  \bq = 1 \:,
 \label{eq:robust213}
\end{eqnarray}
lies in $\innov(\calS_1)$. Therefore, iPursuit exploits this result, combined with the fact that $\innov(\calS_i)$ is orthogonal to $\oplus_{i=2}^N \calS_i$ per Definition \ref{def:innv_subs}, to directly separate out the different subspaces successively. In  contrast, DSC is a spectral-clustering-based approach which uses the outcome of direction search to build a similarity matrix. The main restriction of iPursuit is that it requires every subspace to carry innovation relative to the other subspaces. In other words, iPursuit \emph{requires that no subspace lies in the direct sum of the other subspaces}.
DSC does not have such restrictions. For illustration, the first row of Fig. \ref{fig:example} indeed shows the orthogonality of the optimal direction to $\oplus_{i=2}^4 \calS_i$ when $M_1 = 40$. However, when $M_1 = 20$ in the last row of Fig. \ref{fig:example}, the requirement of iPursuit is violated and iPursuit cannot yield correct clustering. On the other hand, DSC samples few columns corresponding to the largest elements of $|\ba^T \bX|$, which all lie in the first cluster.
Therefore, DSC can form a proper neighborhood set even if the subspaces do not have relative innovations.


\begin{figure}[t!]
	\centering
    \includegraphics[width=0.5\textwidth]{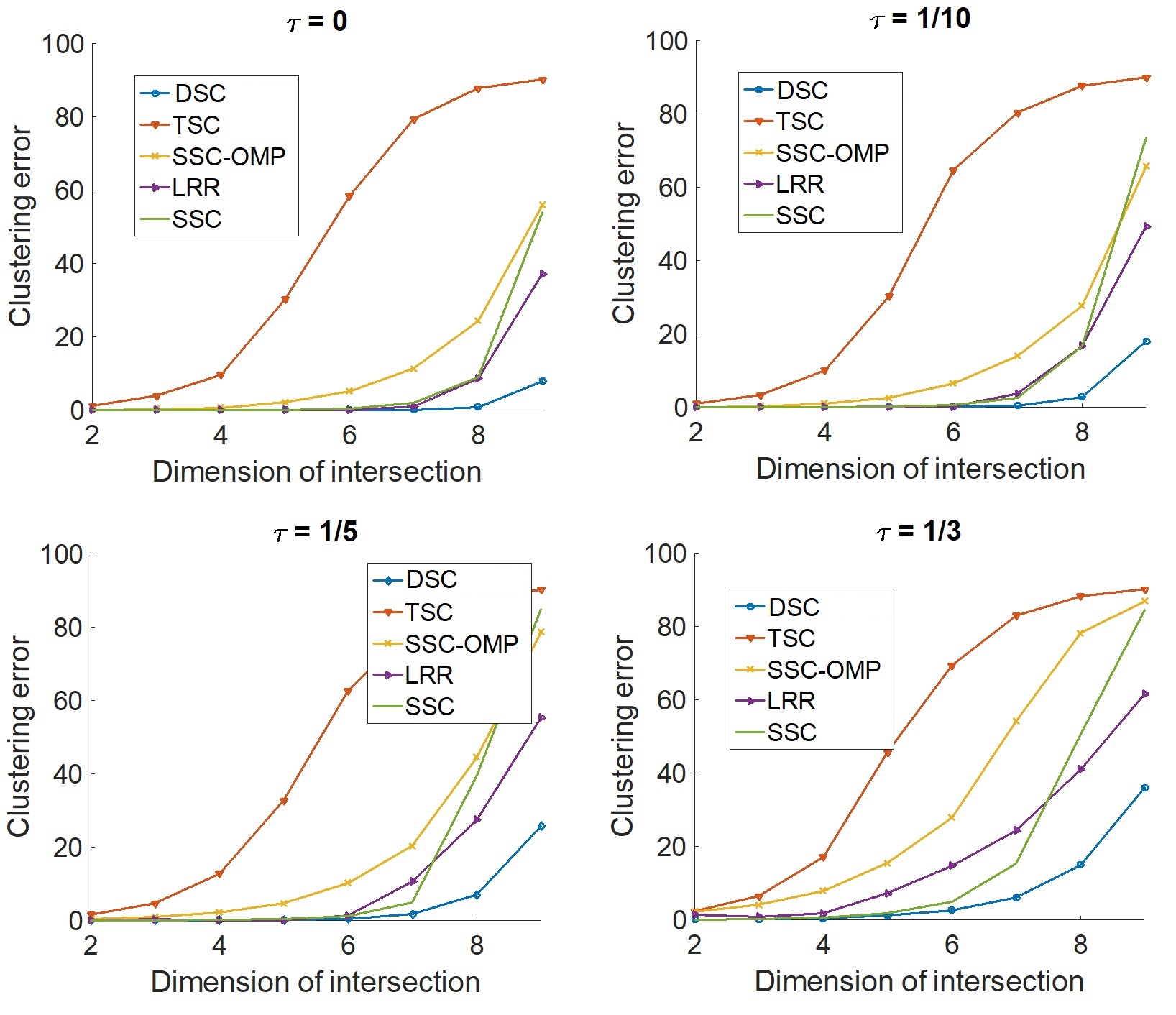}
    \vspace{-.3in}
    \caption{  Performance of the algorithms versus the dimension of intersection
for different noise levels.}
    \label{fig:noiseee}
\end{figure}

\subsection{Solving the proposed optimization problem}
In this section, we use an Alternating Direction Method of
Multipliers (ADMM) \cite{boyd2011distributed} to develop an efficient algorithm for
solving (\ref{eq: sparse rep}) which is a generalized form of (\ref{eq:main_op_p}).
The optimization problem (\ref{eq: sparse rep}) is equivalent to
\begin{eqnarray}
\begin{aligned}
& \underset{\bA , \bZ}{\min}
& & \| \bT \|_{1,p} + \gamma \|\bZ\|_1  \\
& \text{subject to}
& & \bA = \bX \: \bU , \: \bU = \bZ \\
& & &  \bT = \bX^T \bA , \: \: \: \text{and} \: \: \: \text{diag} \left( \bA^T \bX\right)  = \textbf{1} \:.
\end{aligned}
\label{eq:eqv}
\end{eqnarray}
 The Lagrangian function of
(\ref{eq:eqv}) can be written as
\begin{eqnarray}
\begin{aligned}
&\calL(\bT , \bA, \bU, \bZ) = \| \bT \|_{1,p} + \gamma \| \bZ \|_1 + \frac{\mu}{2} \| \bA - \bX \bU \|_F^2 + \\
& \frac{\mu}{2} \left( \text{diag} \left( \bA^T \bX\right)  - \textbf{1} \right)_2^2 + \frac{\mu}{2} \| \bT - \bX^T \bA\|_F^2 + \frac{\mu}{2} \|\bZ - \bU \|_F^2  \\
& + \text{tr} \left( \bY_1^T (\bA - \bX\bU) \right) + \text{tr} \left( \by_2^T \left(\text{diag} \left( \bA^T \bX\right)  - \textbf{1} \right) \right) \\
& + \text{tr} \left( \bY_3^T (\bT  - \bX^T \bA) \right) + \text{tr} \left( \bY_4^T (\bZ  - \bU) \right) \:,
\end{aligned}
\label{eq:lagr}
\end{eqnarray}
where $\mu$ is the regularization parameter.
The ADMM approach is an iterative procedure. Define $\left( \bA_k,\bU_k,\bZ_k, \bT_k \right)$
as the optimization variables and
$\left( \bY_1^k,\by_2^k, \bY_3^k, \bY_4^k  \right)$ as the Lagrange multipliers at the $k^{\text{th}}$ iteration. Define $\bG_1 = \mu^{-1} ( \bI + 2 \bX \bX^T )^{-1}$, $\bG_2 = \mu^{-1} ( \bI +  \bX^T \bX )^{-1}$,  and define the element-wise function $\calT_{\epsilon}(c)$ as $\calT_{\epsilon}(c) = \sgn(c) \max( |c| - \epsilon , 0)$. Define a column-wise  operator $\bH = \calC_{\epsilon} (\bC)$ as follows: set
$\bh_i$ equal to zero  if $\| \bc_i\|_2 \leq \epsilon$, otherwise set $\bh_i = \bc_i - \epsilon \: \bc_i/ \|\bc_i\|_2 $.
Each iteration consists of the following steps:
\begin{eqnarray}
\begin{aligned}
&\bA_{k+1} = \bG_1 \big(\mu \bX \bU_k + \mu \bX  + \mu \bX \bT_k -  \bY_1^k \\
& \quad \quad\qquad\qquad - \bX \: \left( \text{diag}(\by_2^k) \right) + \bX \bY_3^k \big) \\
& \text{if} \: p=1 : \bT_{k+1} = \calT_{\mu^{-1}} (\bX^T \bA_{k+1} - \mu^{-1} \bY_3^k)\\
& \text{if} \: p=2 : \bT_{k+1} = \calC_{\mu^{-1}} (\bX^T \bA_{k+1} - \mu^{-1} \bY_3^k) \\
& \bZ_{k+1} = \calT_{\mu^{-1} \gamma} (\bU_k - \mu^{-1} \bY_4^k) \\
& \bU_{k+1} = \bG_2 \left(\mu \bX^T \bA_{k+1} + \mu \bZ_{k+1} + \bX^T \bY_1^k + \bY_4^k \right) \\
& \bY1^{k+1} = \bY_1^k + \mu (\bA_{k+1} - \bX \bU_{k+1})\\
& \by_2^{k+1} = \by_2^{k} + \mu  \left(\text{diag} \left( \bA^T_{k+1} \bX\right)  - \textbf{1} \right) \\
& \bY_3^{k+1} = \bY_3^k + \mu (\bT_{k+1}  - \bX^T \bA_{k+1}) \\
& \bY_4^{k+1} = \bY_4^k + \mu (\bZ_{k+1} - \bU_{k+1})   \: .
\end{aligned}
\end{eqnarray}
These steps are repeated until the algorithm converges or
the number of iterations exceeds a predefined threshold.

The complexity of the initialization step (obtaining the matrices $\bG_1$ and $\bG_2$) is roughly $\calO(M_2^2 r)$. The order of complexity of each iteration is also $\calO(r M_2^2)$. Thus, the overall complexity is $\calO(T M_2^2 r + M_2 M_1 r)$, where $T$ is the number of iterations and the second term corresponds to the complexity of calculating the matrix $\bX$.


\section{Numerical Simulations}
In this section, we study the performance of DSC with both synthetic and real data.
In the experiments with synthetic data, the data lies in a union of subspaces $\{ \calS_i \}_{i=1}^N$ where $
\calS_i = \calM \oplus \calR_i \:
$. The subspace $\calM$ is a random $y$-dimensional subspace and $\{ \calS_i \}_{i=1}^N$ are random $d$-dimensional subspaces. Hence, the dimension of the intersection between the subspaces is equal to $y$.
The data points are distributed uniformly at random within the subspaces, i.e., a data point lying in an $r_i$-dimensional subspace $\calS_i$ is generated as $\bV_i \bg$, where the elements of $\bg \in \mathbb{R}^{r_i}$ are sampled independently from a standard normal distribution $\calN(0,1)$ and $\bV_i$ is an orthonormal basis for $\calS_i$.
If $\bM_2^{'}$ is the number of misclassified data points, the clustering error is defined as $100\times \frac{M_2^{'}}{M_2}$.
DSC is compared against SSC \cite{elhamifar2013sparse}, LRR \cite{liu2013robust}, TSC \cite{heckel2013robust}, SSC-OMP \cite{dyer2013greedy}, and SCC \cite{chen2009spectral}.
In the simulations with synthetic data, the performance of DSC with $p=1$ and $p=2$ are similar. However, in the face clustering example, DSC yields better perfromance with $p=2$. Thus, we report all the results with $p=2$. In all experiments, $\mu=3.3$ and $\gamma=0.01$.

\subsection{Noisy data}
In this section, we study the performance of DSC with noisy data. The data points lie in a union of 20 10-dimensional linear subspaces and $M_1 = 40$. There are 100 data points in each cluster.
 The noisy data matrix $\bD_e \in \mathbb{R}^{40 \times 2000}$ is obtained as $\bD_e=\bD + \alpha \bE$, where $\bD $ follows Data Model 1, the elements of $\bE$ are sampled from $\calN (0,1)$, and $\alpha$ determines the relative power of the noise component.
Fig. \ref{fig:noiseee} shows the performance of the different algorithms versus the dimension of  intersection for
$
\tau := \frac{\| \alpha \bE \|_F}{\| \bD \|_F}
$
equal to $0$, $1/10$, $1/5$ and $1/3$.
It is worth noting that in this experiment not all subspaces have relative innovations, which excludes iPursuit as a feasible choice. As shown, the proposed approach notably outperforms the other spectral-clustering-based algorithms in all four scenarios.

\begin{figure}[t!]
	\centering
    \includegraphics[width=0.4\textwidth]{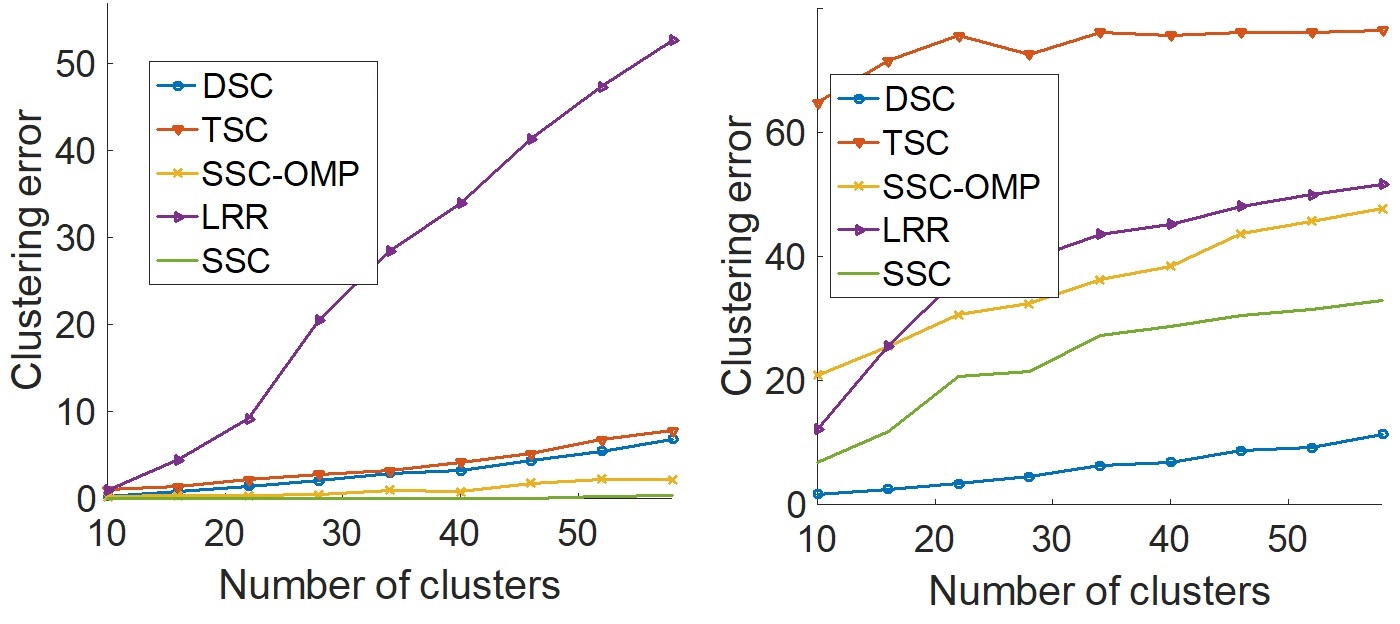}
    \vspace{-.13in}
    \caption{Performance with different number of data clusters. \textbf{Left:} the dimension of intersection $y=0$, \textbf{Right:} $y=4$.}
    \label{fig:NN}
\end{figure}

\subsection{Clustering error versus $N$}
Here, we investigate the performance of the algorithms when there is a large number of clusters. The data follows Data Model 1, the dimension of each subspace is equal to 6, and $M_1 = 20$. There are 60 data points in each cluster. Fig. \ref{fig:NN} shows the clustering error versus the number of subspaces.
In the left plot $y=0$ and all algorithms expect for LRR yield accurate clustering.
In the right plot $y=4$, in which case the clustering error of all algorithms except for DSC notably increases with the number of subspaces. 

\subsection{Face clustering}
Face clustering is a challenging and practical application of subspace clustering \cite{elhamifar2013sparse}. We use the Extended Yale B dataset, which contains 64 images for each of 38 individuals in frontal
view and different illumination conditions \cite{lee2005acquiring}. The faces corresponding to each subject can be approximated with a low-dimensional subspace. Thus, a data set containing face images from multiple subjects can be modeled as a union of subspaces.

We apply DSC to  face clustering  and present results for a different number of clusters in Table \ref{tab:faces_result}.
The performance is also compared with SSC, SCC, and TSC. Heretofore, SSC yielded the best known result for this problem. 
For each number of clusters shown (except 38), we ran the algorithms over 50 different random combinations of subjects from the 38 clusters. To expedite the runtime, we project the data on the span of the first 500 left singular vectors, which does not affect the performance of the algorithms (expect SSC). 
For SSC, we report the results without projection (SSC) and with projection (SSC-P). As shown, DSC yields accurate clustering and notably outperforms the performance achieved by SSC.

\begin{table}
\centering
\caption{Clustering error ($\%$) of different algorithms on the Extended Yale B dataset.  }
\begin{tabular}{|c  |c| c | c|c|c| }
\hline
  $\#$ of    &    &   &   &  &  \\
    subjects    &  DSC  &  SSC &  SSC-P & SCC &  TSC \\
    \hline
  5  & \textbf{ 2.56 } & 4.24  & 29.04  & 62.62 & 25.62 \\
\hline
  10  & \textbf{4.88}   &  9.53 & 32.76 & 74.13  & 40.46 \\
\hline
15  & \textbf{4.71}   & 15.66   & 34.21  & 77.02  & 44.79 \\
\hline
 20   & \textbf{6.45}   &  19.95  & 33.67 & 78.50 & 45.30 \\
\hline
 25   & \textbf{8.53}   & 24.76
  & 50.19 & 79.37 & 46.46 \\
  \hline
   38   & \textbf{8.84}   & 27.47
  & 50.37 & 88.86 & 47.12 \\
\hline
\end{tabular}
\label{tab:faces_result}
\end{table}

{
\bibliography{bibfile}}
\bibliographystyle{IEEEtran}


\end{document}